# Dimension-Specific Imbalance and an Adaptive Hybrid Label Strategy for Multi-Task Affective State Recognition in Classroom Video

Xiangqian Li
The Education University of Hong Kong
lixiangqian2003@gmail.com

## Abstract

Recognition of student affective states from classroom video is constrained by a class imbalance problem whose true nature is, we argue, under-analyzed. We show that on DAiSEE — the de facto benchmark for this task — class imbalance is fundamentally a dimension-specific phenomenon: raw imbalance ratios reach as high as approximately 79:1 (Frustration), and the structure of imbalance — not only its magnitude — varies across dimensions, so uniform algorithmic treatments (aggressive class reweighting, focal loss, uniform binary simplification) fail in dimension-dependent ways. We propose the Adaptive Hybrid Label Strategy (AHLS), which assigns four-level classification to dimensions with manageable imbalance and binary classification to severely long-tailed dimensions, coupled with a null-class placeholder mechanism that stabilizes multi-task optimization by compressing the maximum effective inverse-frequency weight ratio from as high as ~79× to at most ~8.5× across the four dimensions. Validated on a lightweight FERShuffleNetV2 + LTCN architecture (0.39 M parameters, 0.07 G FLOPs), the strategy attains the highest mean Macro-F1 of 47.70% across all four affective dimensions among 11 compared methods, including five state-of-the-art deep models (up to 167× larger) and five traditional machine-learning baselines. A cross-model behavioral analysis on DAiSEE indicates that, in this setting, classification strategy may influence minority-state recognition more strongly than further backbone scaling alone. We also report evidence of an annotation-density limitation of DAiSEE on minority affective states, which suggests that benchmark design may bound future progress as much as algorithmic refinement.

## I. Introduction

Automatic assessment of student engagement [21] and related affective states from classroom video has become a foundational requirement for adaptive learning systems, real-time instructor feedback, and large-scale educational analytics. Among the publicly available benchmarks, the DAiSEE corpus [1] has become the de facto evaluation suite, defining the task as multi-task classification over four affective dimensions — Engagement, Boredom, Confusion, and Frustration — each annotated on a four-level intensity scale. The literature on DAiSEE now spans nearly a decade and reports substantial improvements in mean accuracy through progressively larger architectures, from facial-expression-based classifiers [2] and convolutional-temporal hybrids [3], [5] to 3D video transformers [4].

Yet a persistent observation runs across this literature: while mean accuracy improves, minority-class recall on Confusion and Frustration is rarely reported, and competing methods frequently report results that are difficult to reconcile when reweighting strategies or label-mapping decisions vary [3], [5], [6]. The standard explanation invokes the long-tailed label distribution of DAiSEE [1] and motivates the use of focal loss [7], class-balanced loss [8], or label-distribution-aware margins [9]. We argue that this framing — class imbalance as a single, uniform property of DAiSEE — obscures the empirical structure of the problem and is responsible for two recurrent pathologies in reported results: aggressive reweighting destabilizes optimization on the most long-tailed dimensions, and uniform binary simplification discards information on the moderately imbalanced ones.

This work begins from a quantitative observation: imbalance on DAiSEE is dimension-specific. Engagement and Boredom remain tractable under four-level classification, with effective ratios of approximately 8.5:1 and 1.7:1 after rare-class merging, regimes in which weighted cross-entropy converges stably; Confusion and Frustration, by contrast, are severely long-tailed in raw form (approximately 60:1 and 79:1), where naive four-level reweighting destabilized training in our preliminary runs. A single algorithmic mechanism applied uniformly across dimensions cannot accommodate this heterogeneity without trading recall in one dimension for collapse in another. We therefore propose a per-dimension treatment of classification granularity, which we formalize as the Adaptive Hybrid Label Strategy (AHLS).

We validate the strategy on a deliberately lightweight architecture — FERShuffleNetV2 [10] as a per-frame backbone followed by a Lightweight Temporal Convolutional Network (LTCN) [11] — chosen because edge deployment in classrooms imposes real constraints on parameter count and latency. The validation architecture is not the contribution of this paper; the strategy is. Specifically, we show that the hybrid treatment yields the highest mean Macro-F1 on a 0.39 M parameter model while matching far larger models on Confusion minority recall. To this end, we perform a cross-model behavioral analysis on eleven methods — five state-of-the-art deep models, five traditional machine-learning baselines, and the proposed model — and report per-class recall in addition to aggregate metrics. The pattern that emerges, which we develop in Section VI, is that on DAiSEE the choice of classification strategy appears to influence minority-state recognition more strongly than further backbone scaling alone.

Our contributions are as follows.

1) We quantitatively characterize the dimension-specific imbalance phenomenon in DAiSEE, showing that affective dimensions reach raw imbalance ratios as high as approximately 79:1 (Frustration) and differ markedly in the structure of their imbalance, and we argue that this heterogeneity demands per-dimension methodological treatment rather than uniform reweighting or binary simplification.

2) We propose the Adaptive Hybrid Label Strategy (AHLS), which assigns four-level granularity to dimensions with manageable imbalance and binary granularity to severely long-tailed dimensions, coupled with a null-class placeholder mechanism that stabilizes multi-task optimization by compressing the maximum effective inverse-frequency weight ratio from as high as ~79× to at most ~8.5× across the four dimensions.

3) We provide a cross-model behavioral analysis of class imbalance on DAiSEE, in which a 0.39 M parameter model with the proposed strategy achieves higher mean Macro-F1 than a 64.97 M parameter model trained without it; this suggests that, in this setting, classification strategy can affect balanced minority-state performance more strongly than backbone scaling alone.

4) We provide initial evidence suggesting that progress on minority affective state recognition in DAiSEE may be constrained by annotation density rather than algorithmic capability, motivating future investment in benchmark refinement for the affective-computing community.

The remainder of the paper is organized as follows. Section II reviews prior work on DAiSEE-based affective state recognition, class imbalance in deep learning, and multi-task learning under heterogeneous label structures. Section III presents the dimension-specific imbalance analysis and identifies the failure modes of uniform treatments. Section IV develops AHLS, its null-class placeholder mechanism, and the validation architecture. Section V details the experimental protocol over eleven compared methods. Section VI reports the cross-model behavioral analysis, the annotation-density argument, and per-dimension results. Section VII describes a real-time deployment study. Section VIII concludes.

## II. Related Work

### *A. Engagement and Affective State Recognition on DAiSEE*

DAiSEE [1] established a community standard for evaluating affective state recognition from classroom-style video by providing four-level annotations across four dimensions for 9,068 short clips; unlike multimodal physiological corpora such as AMIGOS [12], it relies on webcam video alone. Subsequent work has progressively scaled the backbone capacity: ResNet-based [24] and SE-augmented CNNs [13], TCN-based temporal models [3], EfficientNet backbones with LSTM and bidirectional LSTM heads [5], [14], and most recently 3D video transformers [4]. While these approaches have demonstrated steady improvements in mean accuracy, minority-class detection has received little direct attention: most prior work reports accuracy only [3], [5], [6]; in our re-implementation (TABLE V), Frustration minority recall stays below 40% for four of the five deep baselines. Our work re-examines the source of this difficulty by treating it as a dimension-specific rather than uniform phenomenon.

### *B. Class Imbalance in Deep Learning*

Class imbalance is pervasive in deep learning [29], particularly on long-tailed benchmarks [30]. Algorithmic responses fall into three broad families: data-level approaches (oversampling, undersampling, synthetic data generation) [15]; algorithmic reweighting (inverse-frequency, effective-number-based [8], or label-distribution-aware margin [9]); and loss-function engineering (focal loss [7], asymmetric loss). These methods share an implicit assumption: that the imbalance is a single property of the task, addressable by a single mechanism applied uniformly across outputs. This assumption is well-suited to single-label classification, but, as we show in Section III, it does not generalize to multi-task settings in which the heads exhibit fundamentally different imbalance structures.

### *C. Multi-Task Learning under Heterogeneous Label Structures*

Multi-task learning [16] traditionally assumes that the per-task losses operate on comparable label spaces, with task-balancing handled by scalar weights or uncertainty-based combination [17]. A small but growing literature considers tasks with structurally heterogeneous outputs (e.g., regression versus classification jointly [18]). To our knowledge, however, no prior work on affective state recognition has explicitly varied the classification granularity per dimension as a function of imbalance structure. The proposed strategy positions itself in this lineage: it treats the choice of label granularity as a per-task design variable subject to per-dimension distributional evidence.

## III. Problem Analysis: Dimension-Specific Imbalance

Existing affective state recognition methods on DAiSEE share an unexamined assumption: that class imbalance, while severe, is a uniform property of the dataset to be addressed by a single algorithmic mechanism — focal loss, class weighting, or label smoothing. We argue this assumption is incorrect, and that its incorrectness helps explain the poor minority-class performance observed across the baselines in TABLE V.

### *A. Quantifying Per-Dimension Imbalance on DAiSEE*

DAiSEE comprises four affective dimensions, each annotated independently with four-level intensity labels (Very Low, Low, High, Very High). TABLE I reports the per-class counts over the 8,925 labelled clips used in this study. Two observations follow.

**TABLE I. Per-Dimension Label Distribution and Imbalance Ratios on DAiSEE**

| Dimension | Class 0 | Class 1 | Class 2 | Class 3 | Raw 4-level ratio | Effective ratio (AHLS) | Severity |
|---|---|---|---|---|---|---|---|
| Engagement | 61 | 455 | 4,422 | 3,987 | ≈72:1 | ≈8.5:1 | manageable |
| Boredom | 3,822 | 2,850 | 1,923 | 330 | ≈12:1 | ≈1.7:1 | manageable |
| Confusion | 5,951 | 2,133 | 741 | 100 | ≈60:1 | ≈2.0:1 | severe |
| Frustration | 6,887 | 1,613 | 338 | 87 | ≈79:1 | ≈3.4:1 | extreme |

*Counts compiled from the DAiSEE annotation release [1]. Raw 4-level ratio denotes the ratio between the most-frequent and least-frequent classes under the original four-level annotation, indicating the magnitude of the underlying imbalance problem. Effective ratio (AHLS) denotes the ratio after the AHLS-prescribed class merger (Engagement and Boredom retained as four-level with a null-class placeholder; Confusion and Frustration binarized) computed from the label counts above. The compression from Raw to Effective is the quantitative motivation for AHLS, further analysed in TABLE IV.*

First, the raw magnitude of imbalance varies roughly sixfold across dimensions (12:1 to 79:1): Engagement and Boredom exhibit raw 4-level ratios of approximately 72:1 and 12:1 respectively, while Confusion and Frustration exhibit ratios of approximately 60:1 and 79:1 — values that motivate algorithmic intervention on all four dimensions. Crucially, however, the structure of imbalance differs: Engagement's dominant classes are the two upper intensity levels (High and Very High account for more than 90% of samples), which under the AHLS-prescribed merger compress to an effective ratio of approximately 8.5:1 — a regime in which conventional weighted cross-entropy converges stably. Boredom admits a similar four-level treatment with an effective ratio of approximately 1.7:1. Confusion and Frustration, in contrast, are dominated by a single absent class (Very Low: 66.7% and 77.2% of clips), with the intensity levels thinning rapidly toward Very High (100 and 87 clips); binarization is therefore necessary, compressing their effective ratios to approximately 2.0:1 and 3.4:1 respectively — well within the regime where bounded reweighting is stable. Second, this heterogeneity propagates into optimization: a uniform inverse-frequency weighting scheme applied to the raw four-level Frustration task assigns a maximum-to-minimum class-weight ratio of approximately 79× (TABLE IV), a regime that destabilized training in our preliminary runs (TABLE IV note).

We stress that the effective ratio is an optimization-oriented quantity, computed after the AHLS label mapping, rather than a property of the original DAiSEE annotation distribution. The raw 4-level ratio describes the dataset; the effective ratio describes what the optimizer actually sees once rare classes are merged or binarized. Reporting both makes the gap between the two — the quantity the proposed mapping is designed to reduce — explicit. Fig. 1 plots the per-dimension class counts on a logarithmic scale, where the long tail on Confusion and Frustration is directly visible.

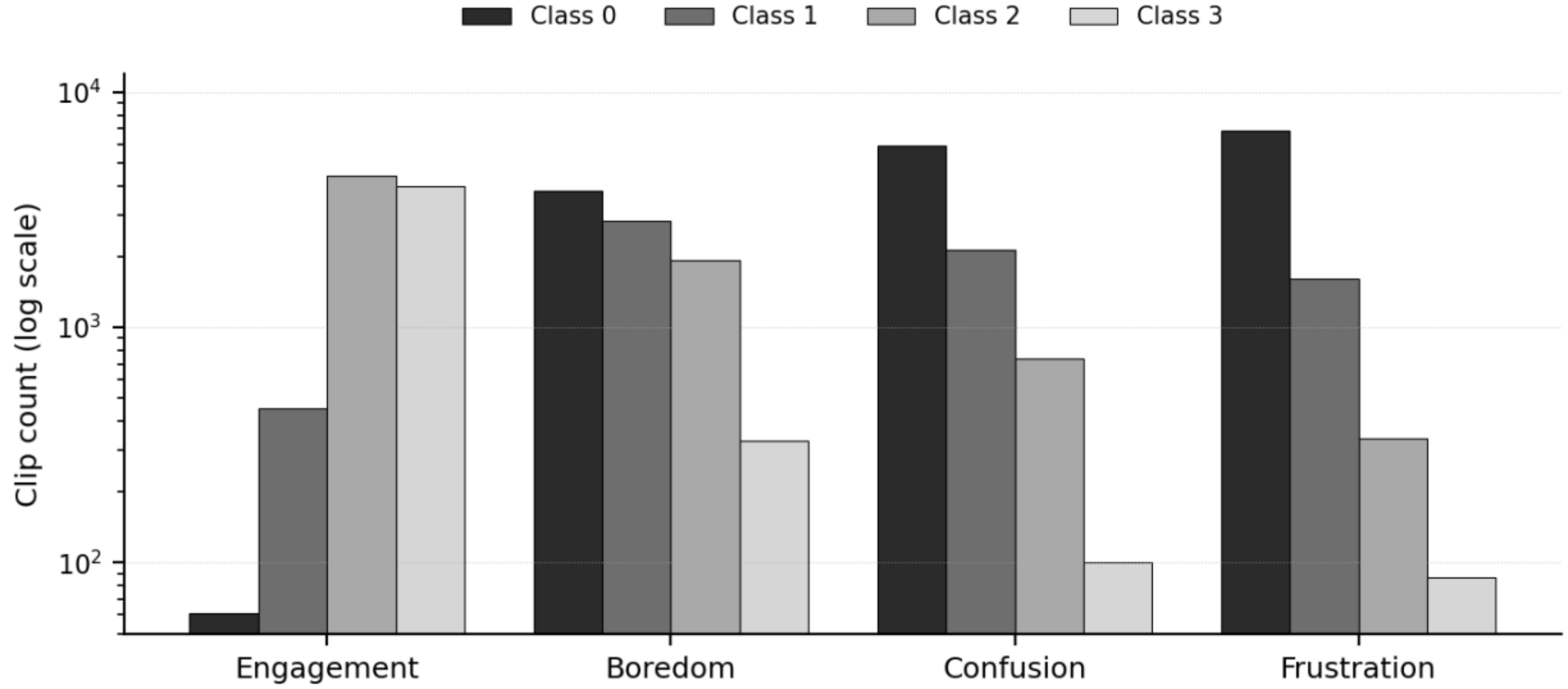

Fig. 1. Per-dimension label distribution on DAiSEE (logarithmic vertical axis). The most-frequent class on Confusion and Frustration exceeds the rarest by roughly two orders of magnitude, whereas on Engagement the raw gap is confined to the two low-intensity levels merged by AHLS, and Boredom is comparatively balanced.

### *B. Failure Modes of Uniform Treatments*

#### *B.1 Full four-level training with standard cross-entropy*

Without adequate reweighting, models on Confusion and Frustration collapse toward the dominant Very-Low class. The cross-model results in TABLE V show this pattern even under the binary formulation: the EfficientNetB7+LSTM baseline, with 167× the parameter count of our proposed model, achieves a Confusion-positive recall of only 42.5% and a Frustration-positive recall of 27.0%. VideoSwinEmotion, despite leading the comparison on mean accuracy, achieves Confusion-positive recall of 10.9% and Frustration-positive recall of 14.8% — a classical majority-class collapse pattern.

#### *B.2 Aggressive class reweighting at the four-level granularity*

Inverse-frequency reweighting on the raw four-level Frustration task yields a maximum-to-minimum class-weight ratio of approximately 79×. In our preliminary runs, effective weight ratios above approximately 10× consistently destabilized optimization (TABLE IV note), in line with the broader observation that inverse-frequency reweighting amplifies noise from rare, possibly mislabelled samples [8], [9]. An analogous over-correction is visible in the inverted recall pattern of the class-balanced ML baselines in TABLE V, where LR — trained with class_weight='balanced' — achieves 62.2% / 62.4% minority recall on Confusion and Frustration at the cost of majority recall collapsing to 38.1% / 41.4%. Mean Macro-F1 drops to 32.93%, the lowest in the comparison: aggressive reweighting trades one collapse for another rather than achieving balanced performance.

#### *B.3 Uniform binary simplification*

An alternative response is to binarize every dimension into positive/negative pairs, sacrificing intensity information for balance. We argue this discards information that Engagement and Boredom can in fact support: once their rarest level is merged into its neighbour, these dimensions yield effective ratios below 10:1 (8.5:1 and 1.7:1) and admit graded classification with conventional weighted loss. Uniform binarization throws away a recoverable signal.

### *C. Design Principles for Heterogeneous Treatment*

Three principles follow from the preceding analysis. First, classification granularity should be a per-dimension design variable determined by the empirical distribution rather than fixed across the multi-task head. Second, when intensity levels are merged for rare-class consolidation, the merger should be implemented in a way that does not break the unified output structure used by the multi-task loss — motivating a null-class placeholder rather than a structural change. Third, per-task loss weighting should be co-designed with the granularity choice so that the effective inverse-frequency weight ratio remains bounded, avoiding the optimization pathologies of aggressive reweighting. These principles are operationalized in Section IV.

## IV. Proposed Method

### *A. Adaptive Hybrid Label Strategy (AHLS)*

AHLS assigns classification granularity per affective dimension as a function of its empirical label distribution. The decision uses two quantities read directly from TABLE I. First, if a single level (Very Low) holds a clear majority of clips — 66.7% for Confusion and 77.2% for Frustration — the dimension is binarized into absent versus present. Otherwise, as for Engagement and Boredom, whose largest class holds under 50% of clips, four-level granularity is retained and the rarest level is merged into its neighbour so that the effective ratio falls below 10:1, a bound chosen from preliminary runs (TABLE IV note). The full mapping is given in TABLE II.

**TABLE II. Adaptive Hybrid Label Mapping**

| Dimension | Task type | Training-label mapping | Placeholder class |
|---|---|---|---|
| Engagement | 4-level | 0,1 → 1; 2 → 2; 3 → 3 | class 0 ($w_c = 0$) |
| Boredom | 4-level | 0 → 0; 1 → 1; 2,3 → 2 | class 3 ($w_c = 0$) |
| Confusion | binary | 0 → 0; 1,2,3 → 1 | — |
| Frustration | binary | 0 → 0; 1,2,3 → 1 | — |

For the two four-level dimensions, the rarest level is merged into an adjacent level rather than removed; the freed output index is retained as a null-class placeholder whose loss weight is fixed to zero:

$$w_c = 0, \quad \text{for } c \in \{\text{placeholder}\} \qquad (1)$$

This design preserves a unified four-class output structure across the two four-level heads — which simplifies the multi-task batch loss — without contributing any gradient signal from the placeholder. Equivalently, the placeholder acts as a structural identity element for the dimension. The effective inverse-frequency class-weight ratio in our setup compresses from approximately 79× (under naive four-level reweighting on Frustration) to approximately 3.4× (under the binary formulation), and to at most 8.5× across all four dimensions, as quantified in TABLE IV.

**TABLE III. Training Hyperparameters**

| Hyperparameter | Value |
|---|---|
| Epochs | 30 |
| Batch size | 16 |
| Initial learning rate | $3 \times 10^{-5}$ |
| Optimizer | AdamW |
| Weight decay | 0.05 |
| LR schedule | Cosine annealing, warm-up 3 epochs |
| Label smoothing $\varepsilon$ | 0.05 |
| Focal loss $\gamma$ (binary) | 2.0 |
| Dropout (heads) | 0.5 |
| Mixed precision | FP16 forward, FP32 master |
| Random seed | 42 (PyTorch / NumPy / sklearn / XGBoost) |
| Model selection criterion | val_score = 0.6 × avg_acc + 0.4 × avg_F1 |

TABLE IV. Compression of Effective Class Weights under AHLS

| Dimension | Granularity | Max/min $w_c$ ratio (raw) | Max/min $w_c$ ratio (AHLS) | Compression factor |
|---|---|---|---|---|
| Engagement | 4-level | ≈ 72.4× | ≈ 8.5× | ≈ 8.5× |
| Boredom | 4-level | ≈ 11.6× | ≈ 1.7× | ≈ 6.8× |
| Confusion | binary | ≈ 59.5× | ≈ 2.0× | ≈ 29.8× |
| Frustration | binary | ≈ 79.2× | ≈ 3.4× | ≈ 23.3× |

*Raw column refers to inverse-frequency weighting under the unmodified four-level distribution; AHLS column refers to inverse-frequency weighting on the post-mapping label space. All values are computed from the label counts in TABLE I. This design choice is motivated by the empirical observation that effective weight ratios above approximately 10× consistently destabilized our optimization in preliminary runs.*

### *B. Validation Architecture: FERShuffleNetV2 + LTCN*

The proposed strategy is, in principle, architecture-agnostic. To validate it under realistic deployment constraints, we use an intentionally lightweight model in the spirit of mobile CNNs [31]: ShuffleNetV2 [10] pre-trained on FER-2013 [19] as a per-frame facial-emotion backbone, followed by a Lightweight Temporal Convolutional Network (LTCN) [11] over a uniformly sampled ten-frame sequence per clip. The architecture totals 0.39 M parameters and 0.07 G FLOPs; training details are reported in Section V-C and TABLE III. We emphasize that this architecture is a means to test the strategy, not the contribution of this paper; whether the same treatment transfers to heavier backbones is left for future work.

The per-frame backbone outputs a seven-dimensional facial-emotion softmax. The three categories with low true-positive rates on the FER backbone (Angry, Fear, Sad) are removed, yielding a four-dimensional encoding $f_t \in \mathbb{R}^4$ for each of $T = 10$ frames. The LTCN consists of residual blocks with causal dilated convolutions (base width 64, kernel size 3, dilation factors {1, 2, 4}); the temporal output is global-average pooled and projected by four task-specific heads. Modelling the temporal evolution of per-frame expression features, rather than single frames, follows spatio-temporal facial-expression recognition practice [22].

### *C. Multi-Task Loss with Hybrid Granularity*

The four-level heads (Engagement, Boredom) use a cross-entropy loss with inverse-frequency weighting and label smoothing:

$$L_4 = -\sum_c w_c \cdot [(1-\varepsilon)\, y_c + \varepsilon / C] \cdot \log p_c,\quad \varepsilon = 0.05 \qquad (2)$$

where $w_c = N / (C \cdot n_c)$ is the inverse-frequency class weight (N total samples, C trainable classes, $n_c$ per-class count), and $\varepsilon = 0.05$ is the label-smoothing coefficient. For the placeholder class, $w_c = 0$ as in (1). The binary heads (Confusion, Frustration) use a class-weighted focal loss:

$$L_2 = -\sum_c w_c \cdot (1 - p_c)^\gamma \cdot \log p_c,\quad \gamma = 2.0 \qquad (3)$$

The total multi-task loss is the weighted sum:

$$L = \lambda_{Eng} \cdot L_{Eng} + \lambda_{Bor} \cdot L_{Bor} + \lambda_{Con} \cdot L_{Con} + \lambda_{Fru} \cdot L_{Fru} \qquad (4)$$

with $\lambda_{Eng} = 0.5$, $\lambda_{Bor} = 1.0$, $\lambda_{Con} = 2.0$, $\lambda_{Fru} = 2.5$. The higher λ values on the binarized dimensions compensate for the lower intrinsic loss magnitude of two-class softmax relative to four-class softmax. This design choice is motivated by the observation that uniform λ produced under-training of the binary heads in preliminary validation runs.

### *D. Conceptual Summary*

Fig. 2 illustrates the validation architecture. Inputs are uniformly sampled clips of ten frames; each frame is processed by the FER-pre-trained ShuffleNetV2 backbone; the four-dimensional per-frame embeddings are passed to the LTCN; four task heads emit predictions under the AHLS mapping. The full pipeline is trained end-to-end with the loss in (4).

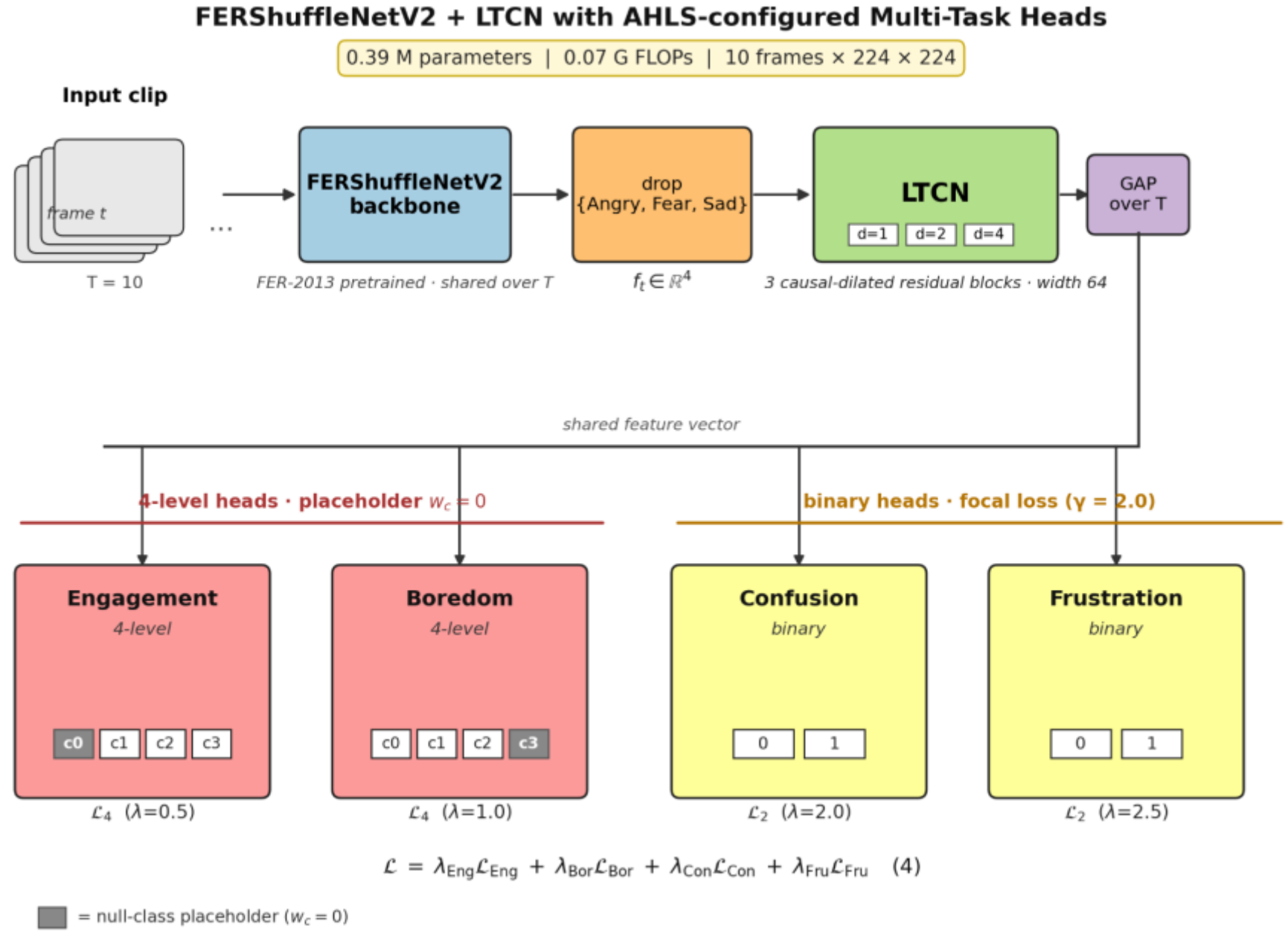

Fig. 2. Validation architecture: FERShuffleNetV2 + LTCN with AHLS-configured multi-task heads.

## V. Experimental Setup

### *A. Dataset and Splits*

All experiments use the DAiSEE corpus [1], comprising 9,068 video clips of approximately ten seconds each, annotated independently for Engagement, Boredom, Confusion, and Frustration on a four-level intensity scale. We use the standard subject-disjoint partition: 5,358 clips for training, 1,429 for validation, and 1,638 for test. All reported results are on the 1,638-clip test set; the validation set is used exclusively for model selection. Per-class counts are given in TABLE I.

### *B. Comparison Methods*

We compare eleven methods spanning state-of-the-art deep learning, traditional machine learning, and the proposed AHLS-trained model.

Deep learning baselines (five). SE-ResNet50+LSTM [13] combines an SE-augmented backbone with a bidirectional LSTM temporal head. ResNet18+TCN [3] uses a ResNet-18 backbone with a temporal convolutional network. OptShuffleNetV2 [10], [20] is an optimized lightweight CNN with a pooling-based temporal aggregator. EfficientNetB7+LSTM [5], [14] uses a 600 × 600 input resolution and scales backbone capacity to 64.97 M parameters. VideoSwinEmotion is our implementation of a Video Swin Transformer [4] adapted with a multi-head classifier for the four DAiSEE dimensions.

Machine-learning baselines (five). K-Nearest Neighbors, XGBoost [23], multi-layer perceptron (MLP), support vector machine (SVM), and logistic regression (LR) are trained on the 40-dimensional feature vectors produced by the FERShuffleNetV2 backbone (ten frames × four emotion channels). All ML baselines use class_weight='balanced' or sample reweighting to counteract imbalance; their hyperparameters are tuned by stratified five-fold cross-validation on the training set. ML inference is performed on CPU.

Proposed model. FERShuffleNetV2 + LTCN trained with AHLS and the loss in (2)–(4). We report the FP32 model (0.39 M parameters, 0.07 G FLOPs). All eleven methods share the same train/val/test partition and the same random seed (42) for reproducibility.

### *C. Implementation and Training Protocol*

Deep models are implemented in PyTorch 2.1 and trained on an NVIDIA A100-SXM4-40GB. Training hyperparameters for the proposed model are listed in TABLE III; optimization uses AdamW [26], [27]. Model selection follows val_score = 0.6 × avg_acc + 0.4 × avg_F1, computed on the 1,429-clip validation set; this scoring rule slightly favours accuracy to avoid rewarding degenerate high-F1 / low-accuracy collapses. The best checkpoint by val_score is evaluated on the test set.

### *D. Evaluation Metrics*

For each dimension we report Accuracy, Macro-F1, and per-class Recall. Across dimensions we report the mean of each. Because model rankings can change with the choice of evaluation metric [28], we fix the primary metric in advance. Macro-F1 is our primary metric because it is the standard scalar summary that remains informative under imbalance: a model that collapses to the majority class on Confusion or Frustration will achieve high accuracy but low Macro-F1. We additionally report per-class Recall for the minority class on Confusion and Frustration (denoted cls1) because this is the operationally meaningful quantity for classroom deployment — false negatives on these states translate directly into missed instructor-intervention opportunities.

## VI. Results and Analysis

### *A. Cross-Model Collapse Analysis: The Natural Ablation*

TABLE V reports per-class recall for all eleven compared methods on Confusion-cls1 and Frustration-cls1 — the two minority classes on the dimensions exhibiting severe imbalance. Three patterns emerge.

**TABLE V. Cross-Model Performance with Per-Class Recall on DAiSEE Test Set**

| Model | Params (M) | Avg Acc (%) | Mean Macro-F1 (%) | Con-cls0 R (%) | Con-cls1 R (%) | Fru-cls0 R (%) | Fru-cls1 R (%) |
|---|---|---|---|---|---|---|---|
| SE-ResNet50+LSTM | 25.26 | 53.72 | 45.58 | 65.6 | 43.9 | 83.6 | 24.8 |
| ResNet18+TCN | 11.66 | 48.86 | 44.17 | 58.1 | 53.9 | 67.5 | 52.1 |
| OptShuffleNetV2 | 0.485 | 49.39 | 42.26 | 68.9 | 37.4 | 72.3 | 36.8 |
| EfficientNetB7+LSTM | 64.97 | 46.08 | 39.26 | 67.6 | 42.5 | 75.4 | 27.0 |
| VideoSwinEmotion | 27.86 | 56.68 | 43.92 | 94.5 | 10.9 | 94.4 | 14.8 |
| KNN | — | 47.53 | 37.07 | 62.5 | 39.4 | 74.4 | 27.6 |
| XGBoost | — | 46.90 | 34.69 | 66.6 | 31.2 | 78.0 | 22.0 |
| MLP | — | 41.74 | 34.60 | 46.7 | 54.3 | 44.0 | 62.4 |
| SVM | — | 41.18 | 33.83 | 44.6 | 55.3 | 51.3 | 50.7 |
| LR | — | 38.86 | 32.93 | 38.1 | 62.2 | 41.4 | 62.4 |
| **Proposed (FERShuffleNetV2 + LTCN, AHLS)** | **0.39** | **48.15** | **47.70** | **52.8** | **42.9** | **82.3** | **15.3** |

*Con-cls1 / Fru-cls1 refer to the minority (positive-affect) class in the binary formulation. ML baselines use class_weight='balanced' or equivalent reweighting; the inversion of the cls0 / cls1 recall pattern in MLP, SVM, and LR reflects an over-correction discussed in the text. Bold-faced row indicates the proposed model.*

First, capacity does not rescue minority-class detection. EfficientNetB7+LSTM (64.97 M parameters, 167× the size of our model) attains 42.5% recall on Confusion-cls1 and 27.0% on Frustration-cls1 — comparable to our 42.9% on Confusion and higher than our 15.3% on Frustration, but with two orders of magnitude more parameters. VideoSwinEmotion (27.86 M parameters, 71× larger) performs strictly worse on the minorities: 10.9% and 14.8% on Confusion-cls1 and Frustration-cls1 respectively, despite achieving the highest mean Accuracy (56.68%) in the comparison. Its Accuracy lead is purchased by 94.5% / 94.4% recall on the corresponding majority classes — the classical signature of majority-class collapse.

Second, classical machine-learning baselines (KNN, XGBoost, MLP, SVM, LR) display a complementary failure mode. Trained with class_weight='balanced' to counteract imbalance, three of them (MLP, SVM, LR) over-correct: minority-class recalls reach 51–62% on Confusion and Frustration (LR achieves 62.2% / 62.4%), but majority-class recalls fall to 38–51% in the same models, while KNN and XGBoost remain majority-leaning, yielding low Macro-F1 due to majority-class false positives. Macro-F1 captures this trade-off: LR, with the highest minority recall in the table, has the lowest mean Macro-F1 (32.93%).

Third, the proposed model attains the highest mean Macro-F1 of 47.70% while avoiding both failure modes on Confusion: its minority recall (42.9%) matches that of models up to 167× larger, without the majority-class collapse of VideoSwinEmotion or the inversion of the balanced ML baselines. Its Frustration-cls1 recall (15.3%) remains low, and ResNet18+TCN achieves a more even cls0/cls1 split on both binary dimensions, so the result is best read as an association between the per-dimension granularity and balanced performance rather than as proof that architecture is irrelevant. The association is at least consistent across the compared model families (CNN + RNN, CNN + TCN, 3D transformer), parameter scales (0.39 M to 64.97 M), and training paradigms (cross-entropy with weighting, focal loss, balanced classical ML) evaluated here.

### *B. Macro-F1 vs. Accuracy: Why the Metric Matters*

The Accuracy-versus-Macro-F1 dissociation is particularly visible on VideoSwinEmotion. Its mean Accuracy of 56.68% leads the comparison by approximately three percentage points, yet its mean Macro-F1 of 43.92% is lower than that of our 0.39 M-parameter model by 3.78 percentage points. Decomposing the Confusion result, the model attains 68.86% Accuracy by predicting the majority class on 94.5% of majority-class samples and on a substantial fraction of minority-class samples — yielding the 10.9% minority recall observed. The Macro-F1 metric penalises this asymmetry, while Accuracy rewards it. We argue that on imbalanced affective dimensions Macro-F1 is the operationally meaningful metric, and Accuracy alone should not be used to declare progress.

### *C. Per-Dimension Macro-F1 Comparison*

**TABLE VI. Per-Dimension Macro-F1 (%) on DAiSEE Test Set**

| Model | Engagement | Boredom | Confusion | Frustration | Mean |
|---|---|---|---|---|---|
| SE-ResNet50+LSTM | 35.27 | 38.40 | 54.30 | 54.36 | 45.58 |
| ResNet18+TCN | 32.04 | 33.63 | 54.26 | 56.73 | 44.17 |
| OptShuffleNetV2 | 33.23 | 29.10 | 53.04 | 53.67 | 42.26 |
| EfficientNetB7+LSTM | 17.75 | 33.45 | 54.73 | 51.10 | 39.26 |
| VideoSwinEmotion | 37.37 | 34.81 | 49.27 | 54.22 | 43.92 |
| KNN | 24.54 | 22.39 | 50.56 | 50.79 | 37.07 |
| XGBoost | 22.64 | 19.33 | 48.91 | 50.02 | 34.69 |
| MLP | 23.55 | 21.38 | 47.74 | 45.72 | 34.60 |
| SVM | 22.81 | 18.96 | 46.83 | 46.70 | 33.83 |
| LR | 22.92 | 19.56 | 45.19 | 44.03 | 32.93 |
| **Proposed (AHLS)** | **29.28** | **33.64** | **46.85** | **48.54** | **47.70** |

*The proposed model does not lead any individual dimension on Macro-F1, but its per-dimension spread (19.3 points) is among the narrowest, comparable to that of SE-ResNet50+LSTM (19.1 points). SE-ResNet50+LSTM leads on Engagement (35.27) and Boredom (38.40); ResNet18+TCN leads on Frustration (56.73); EfficientNetB7+LSTM leads on Confusion (54.73).*

The per-dimension view in TABLE VI reinforces the cross-model argument: most competitors trade performance on one dimension to gain on another. EfficientNetB7+LSTM spans 37 points (17.75 on Engagement to 54.73 on Confusion), and ResNet18+TCN and OptShuffleNetV2 each span about 25 points. The proposed model's spread (19.3 points) is comparable to that of the most even baseline, SE-ResNet50+LSTM (19.1 points), despite a 65× smaller parameter budget. The result is not the highest per-dimension Macro-F1 anywhere, but the highest mean — which we contend is the property that matters for a multi-task system that must operate on all four dimensions concurrently.

### *D. Efficiency Analysis*

**TABLE VII. Efficiency Comparison (single-sample inference)**

| Model | Params (M) | FLOPs (G) | Latency (ms) | FPS |
|---|---|---|---|---|
| EfficientNetB7+LSTM | 64.97 | 390.28 | 94.6 | 11 |
| VideoSwinEmotion | 27.86 | 14.96 | 8.7 | 114 |
| SE-ResNet50+LSTM | 25.26 | 41.35 | 5.0 | 201 |
| ResNet18+TCN | 11.66 | 18.24 | 4.6 | 226 |
| OptShuffleNetV2 | 0.485 | 0.44 | 4.4 | 228 |
| **Proposed (AHLS)** | **0.39** | **0.07** | **5.4** | **187** |
| KNN (CPU) | — | — | 11.7 | 86 |
| XGBoost (CPU) | — | — | 1.5 | 658 |
| SVM (CPU) | — | — | 1.3 | 770 |
| LR (CPU) | — | — | 0.6 | 1689 |

*Deep-learning latencies measured on NVIDIA A100, batch size 1; machine-learning baselines measured on CPU and are not directly comparable. The proposed model is the smallest deep-learning entry by both parameter count and FLOPs, followed by OptShuffleNetV2 (0.485 M).*

On efficiency, the proposed model occupies a favourable corner of the parameter / accuracy plane: 0.39 M parameters and 0.07 G FLOPs, with 5.4 ms single-sample latency on an A100 GPU (187 FPS). This places it at approximately 1/71 the parameter count of VideoSwinEmotion and 1/167 that of EfficientNetB7+LSTM, while leading both on mean Macro-F1. We do not, however, claim Pareto-optimality across all four dimensions: VideoSwinEmotion achieves higher mean Accuracy at the cost of minority recall, and deployment scenarios that prioritise mean Accuracy over Macro-F1 may legitimately prefer it. Fig. 3 visualises this trade-off.

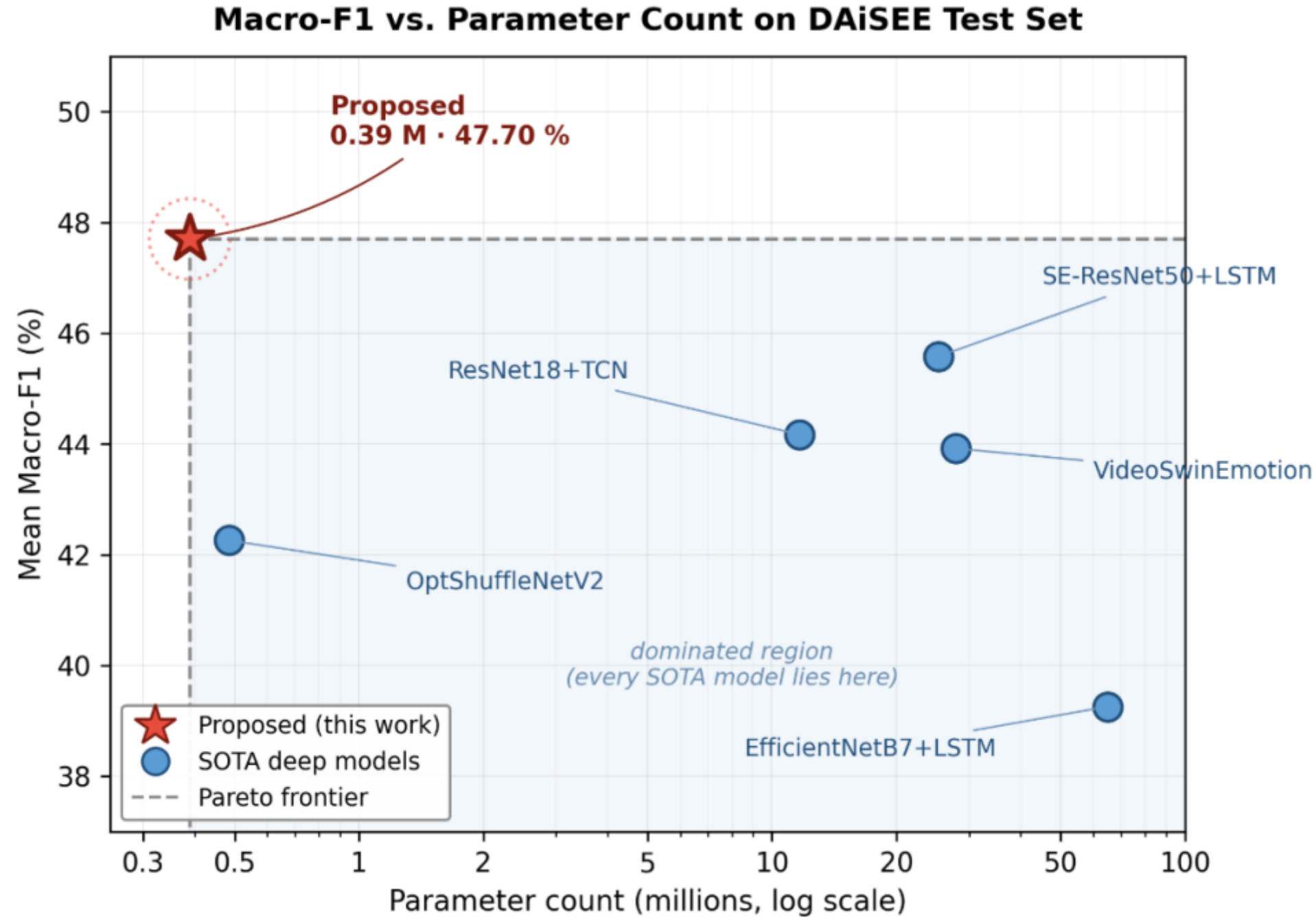

Fig. 3. Mean Macro-F1 versus parameter count for all eleven compared methods on DAiSEE.

### *E. The Fundamental Limitation: Annotation-Density Argument*

We turn now to a question that the cross-model analysis raises but does not resolve: why do all eleven methods — including the proposed model — plateau at modest minority-class recall on Confusion and Frustration?

We argue the answer is annotation-density limitation, not algorithmic limitation. Binarization yields a workable positive class (about one-third of clips for Confusion and just under a quarter for Frustration, TABLE I), but that class is dominated by the Low level; the High and Very High levels that carry the clearest affective signal are sparse — 841 and 425 clips across the whole corpus, and only 100 and 87 at Very High (about 1% each). For supervised learning, this sample density is insufficient to characterise the within-class manifold of these affective states, particularly given the moderate inter-rater agreement reported in the DAiSEE annotation protocol on Frustration in particular [1]. The 167× capacity escalation from our model to EfficientNetB7+LSTM produces no gain on Confusion-cls1 (42.5% versus our 42.9%) and, on Frustration-cls1, a gain (27.0% versus our 15.3%) far smaller than the capacity difference would suggest. More broadly, none of the eleven methods exceeds roughly 55% minority recall while keeping majority recall above 50%, a pattern consistent with a data-bound rather than capacity-bound problem. We also note that the ML baselines (LR, MLP) achieve higher minority recall only by inverting the imbalance — trading majority recall for minority recall — which does not constitute genuine progress on the underlying recognition problem.

This argument has two implications. For methods research, it suggests that further algorithmic refinement appears to yield diminishing returns on DAiSEE minority-state detection: the compared models span more than two orders of magnitude in backbone capacity without proportional minority-recall improvement. For benchmark design, it suggests value in investing in larger and more class-balanced affective video datasets, ideally with multi-annotator consensus labelling on minority states. These observations suggest that future gains on DAiSEE-style minority affective states may increasingly depend on dataset quality — particularly annotation density and inter-rater consensus on rare classes — rather than on architectural scaling alone. We acknowledge that this framing shifts focus away from method-level competition toward dataset-level investment, which we believe is the conclusion best supported by the evidence in TABLE V.

## VII. Deployment Case Study

### *A. Real-Time Classroom Pipeline*

To validate the operational utility of dimension-specific affective modelling in a realistic classroom setting, we deployed a real-time video-analysis pipeline in a controlled instructional environment. The pipeline consists of: (i) YOLOv8 [25] face detection at 30 FPS; (ii) an IoU-based per-student tracker that maintains identity across frames; (iii) a clip-buffer that accumulates ten-frame windows per tracked face; and (iv) a per-clip affective-state classifier that emits the four-dimension output. The pipeline runs on a single consumer-grade workstation (NVIDIA Titan V); Fig. 4 shows the pipeline schematic.

### *B. Backbone Choice for the Deployed System*

We note transparently that the system used in the deployment study employs a VideoSwinEmotion backbone for the affective classifier rather than the proposed lightweight FERShuffleNetV2 + LTCN model. The choice reflects an engineering decision made earlier in the project lifecycle, before the present cross-model analysis was complete: VideoSwinEmotion had been integrated and validated for the deployment pipeline on the grounds of mean Accuracy. The subsequent dimension-specific analysis reported in Section VI shows that mean Accuracy is a misleading proxy on this benchmark, and the lightweight proposed model would be the principled choice for deployment scenarios that prioritise minority-state detection. Field validation of the lightweight model in classroom conditions is left for future work; the proposed model's 0.39 M parameter count and 5.4 ms single-sample latency on an A100 — and correspondingly modest hardware requirements on consumer GPUs — make it a stronger candidate for edge deployment in resource-constrained classrooms.

### *C. System-Level Validation*

The deployed pipeline was validated on a ten-student classroom session. End-to-end per-student latency (detection + tracking + ten-frame clip inference) remained below 100 ms throughout the session, supporting real-time operation at the per-clip granularity. Qualitative inspection by an experienced instructor confirmed interpretable trajectories of Engagement and Boredom; Confusion and Frustration trajectories were sparser, consistent with the minority-state observations of Section VI. We emphasise that this is a single-session, small-N validation; quantitative human-rater agreement on the deployment outputs is pending.

### *D. Privacy Considerations*

All face detections in the deployment study were processed in-memory on the local workstation and never transmitted off-device. No identifiable face crops or raw video were stored after inference. Per-student state trajectories were aggregated to anonymous tracker indices for instructor review, with student-identifier mapping remaining under instructor control. The pipeline is compatible with GDPR-style data-minimisation requirements provided the workstation deployment is appropriately access-controlled.

### *E. Edge Deployment Future Work*

Future deployment work will replace the VideoSwinEmotion backbone with the lightweight proposed model, supporting deployment on embedded GPU hardware (e.g., Jetson Orin) where the 0.07 G FLOPs / 5.4 ms latency profile of the proposed model becomes operationally advantageous. Field validation under genuinely heterogeneous classroom conditions — varying illumination, occlusion patterns, and student populations — is the natural next step.

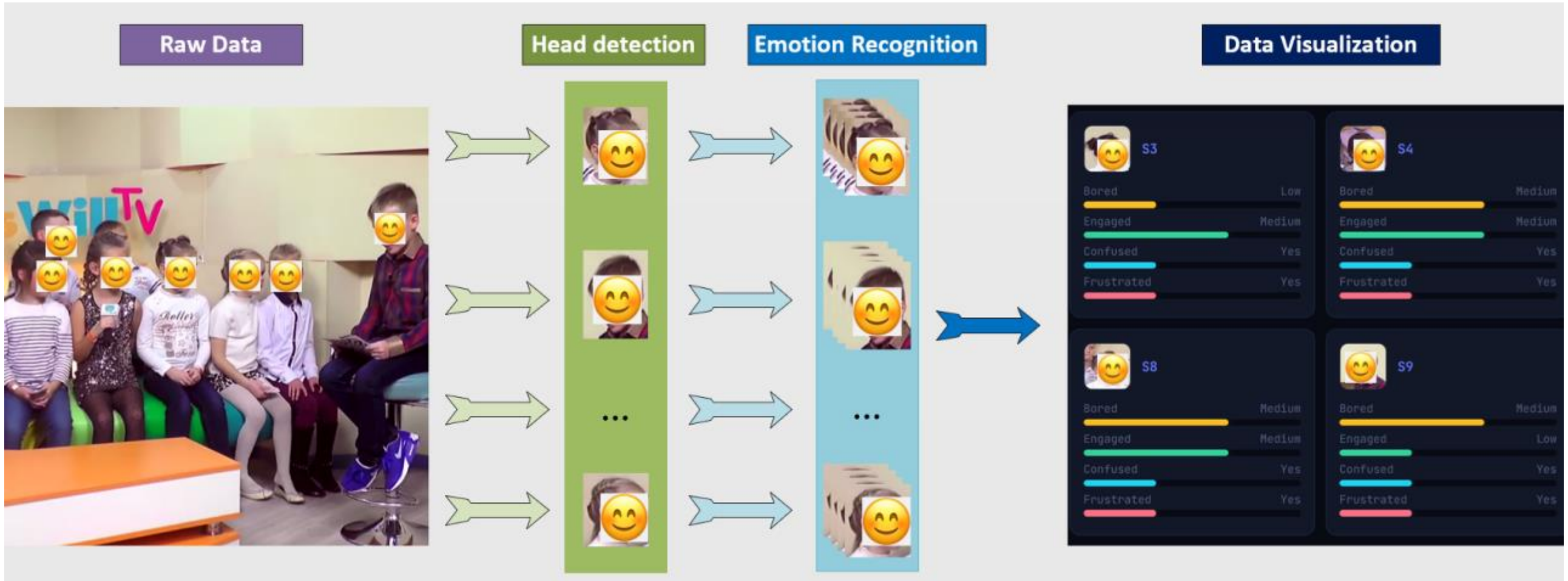

Fig. 4. Real-time deployment pipeline schematic.

## VIII. Discussion and Conclusion

This paper investigated dimension-specific class imbalance in multi-dimensional affective state recognition on DAiSEE. We proposed the Adaptive Hybrid Label Strategy (AHLS), validated it on a 0.39 M parameter architecture, and used cross-model analysis across eleven methods to show that, under the DAiSEE setting, the choice of classification strategy was associated with balanced Macro-F1 more strongly than backbone scale alone. Based on the observation that none of the eleven compared methods exceeds roughly 55% minority-class recall while keeping majority-class recall above 50%, we further argued that progress on DAiSEE may increasingly depend on annotation density rather than on architectural scaling alone.

We acknowledge four limitations. First, our minority-class recall (Confusion 42.9%, Frustration 15.3%) remains modest in absolute terms and, on Frustration, is below that of several baselines; the annotation-density explanation we offer is a hypothesis that requires validation on alternative datasets. Second, all results come from a single training run (seed 42), and we do not report a same-backbone ablation of four-level, binary, and hybrid heads; per-seed variance and this ablation are left for future work. Third, the deployment case study uses a heavier backbone than our proposed model; field validation of the lightweight model in genuine edge environments is pending. Fourth, our findings are DAiSEE-specific, and the generalization of AHLS to other affective video benchmarks requires replication.

Future work will pursue mixed-precision INT8 quantization for edge deployment, knowledge distillation from heavier teacher models [32], and — most importantly — collaboration with the affective-computing community on a successor benchmark addressing the annotation-density limitation identified here.